\documentclass[journal]{IEEEtran}
\usepackage{amsmath,amsfonts}
\usepackage{algorithmic}
\usepackage{array}
\usepackage[ruled,vlined]{algorithm2e}
\usepackage{url}
\usepackage[T1]{fontenc}
\usepackage[utf8]{inputenc}  % For older engines (not needed with LuaLaTeX/XeLaTeX)
\usepackage{xcolor}
\newcommand{\rev}[1]{#1}
\usepackage{graphicx}
\ifCLASSINFOpdf
\else  
\fi

\usepackage{cite}
\usepackage{multirow}
\usepackage{caption}
\usepackage{subcaption}
\begin{document}
	
	\title{Enhancing Strawberry Yield Forecasting with Backcasted IoT Sensor Data and Machine Learning}
	\author{Tewodros~Alemu Ayall$^\dagger$,
		Andy~Li$^\dagger$,
		Matthew~Beddows,
		Milan~Markovic,
		and~Georgios~Leontidis% <-this % stops a space
	%	\thanks{Manuscript received  November 14, 2025 }
		\thanks{TAA, AL, MB, and MM are with The School of Natural and Computing Sciences and the Interdisciplinary Institute at the University of Aberdeen, Aberdeen, AB243FX, UK. GL is with UiT The Arctic University of Norway (e-mails: meettedy2123@gmail.com, [a.li.21, m.beddows.21, milan.markovic]@abdn.ac.uk, georgios.leontidis@uit.no). $\dagger$ co-first authors.}}% 
	
%	\markboth{IEEE TRANSACTIONS ON AGRIFOOD ELECTRONICS, ~Vol.~0, No.~0, August~2025}%
%	{Shell \MakeLowercase{\textit{et al.}}: Enhancing Strawberry Yield Forecasting with Backcasted IoT Sensor Data and Machine Learning}

	\maketitle
	
	\begin{abstract}
		Rapid global population growth underscores the need for digitally enabled agricultural systems that support sustainable food production and data-driven resource management for farmers and stakeholders. The adoption of Internet of Things (IoT) technologies--capable of capturing real-time environmental (e.g., temperature, humidity) and operational (e.g., irrigation) parameters--is a crucial step toward enabling advanced applications such as AI-based yield forecasting. However, the effectiveness of such models is often constrained by limited data availability, particularly in dynamic farm environments where IoT observations must be accumulated over multiple growing seasons.
		In this study, we deployed IoT sensors in strawberry production polytunnels over two growing seasons to collect data on water usage, internal and external temperature and humidity, soil moisture, soil temperature, and photosynthetically active radiation. These observations were combined with manually recorded yield data spanning four seasons. To address gaps in IoT data for the two seasons without sensor coverage, we developed an AI-based backcasting approach that synthesizes missing sensor observations using historical weather data from a nearby station and existing polytunnel measurements. We then trained AI-based yield forecasting models using both real and synthetic datasets. In this retrospective evaluation, results show that incorporating synthetic data improved yield forecasting accuracy, with models trained on the combined dataset outperforming those using only real sensor, weather, and yield data.
	\end{abstract}
	
	
	\IEEEpeerreviewmaketitle

	\section{Introduction}
	\label{sec:sample1}
	\IEEEPARstart{A}{grifood systems} (AFS) include a diverse range of industries that produce, process, distribute, and consume agricultural products. Due to unprecedented population growth across the world, which is expected to reach 9.8 billion by 2050 and 11.2 billion by 2100 \cite{abbate2023digital}, achieving future food security remains a challenging goal \cite{ manikas2023systematic}. Agrifood 4.0 \cite{lezoche2020agri} is expected to revolutionize and improve the sustainability of agricultural production systems, with economic, social, and environmental aspects \cite{ konfo2023recent,bahn2021digitalization}. Similar to Industry 4.0 in manufacturing, the Agrifood 4.0 concept promises to transform traditional farming processes through the use of digital technologies such as AI, privacy-preserving technologies, federated learning, and IoT \cite{lezoche2020agri,durrant2022role,latino2021agriculture,onoufriou2021fully}. 
	
	Soft fruit production is an example of an AFS with a high impact on global economic growth \cite{ konfo2023recent}. Strawberries are a popular fruit and nutritious food source \cite{ giampieri2015strawberry, hernandez2023current} and can grow in open fields and inside polytunnels or greenhouses \cite{kouloumprouka2024opportunities}, depending on the climatic conditions. Controlled environments improve strawberry production by shielding plants from harsh weather, extending the growing season, and allowing better pest management \cite{tang2020effect, kouloumprouka2024opportunities}. The latest analysis by the Food and Agriculture Organization \cite{FAO} on the global strawberry market reveals a constant growth trend. In 2022, the official figure for world strawberry production surpassed 9.57 million tonnes. This indicates a growing popularity and customer desire for organic and sustainable strawberry production.
	
	Crop yield forecasting is a significant challenge in digital agriculture, and numerous prediction models have been proposed \cite{li2024model, botero2025machine}. Although crop yield forecasting models can predict actual yield with reasonable accuracy, further performance improvements are still needed to make such systems useful in real-world settings, where data is scarce \cite{van2020crop}. Accurate forecasting of crop yields is critical to support farmers in informed decision-making, for example, to optimize labor resources by ensuring that a sufficient number of workers are available throughout the harvest seasons \cite{filippi2019approach,chen2019strawberry}. Crop yield forecasting systems typically rely heavily on sensor observations that cover a range of environmental factors such as weather and field measurements \cite{maskey2019weather}. Environmental parameters such as soil moisture,  humidity,  temperature, soil temperature, and light influence crop growth, affecting enzyme activity and photosynthetic plant growth rate \cite{ferrante2018agronomic}. IoT sensors are ideally suited to observe such environmental parameters in real-time \cite{ raj2021survey}, which have seen use with deep learning (DL) strawberry yield forecasting models \cite{onoufriou2023premonition}. However, the performance of such models is often hindered by the lack of adequate historical sensor data \cite{markovic2024embedding, celis2024review}, which is predominantly absent or exceedingly challenging to obtain, primarily due to the complexities of data collection (e.g., remote deployment environment, costs, etc.) and the perceived sensitivity of information \cite{schauberger2020systematic}. 
	
	In our study, we aimed to address this challenge by extending the training datasets containing the real IoT observations with synthetic observations generated using a backcasting approach which offers inexpensive and time-efficient solutions compared to lengthy multi-season sensor deployments, which are rare in real-world polytunnel settings. In addition, such an approach is especially beneficial for farms where crop configuration is frequently changing hence preventing long-term sensor observations of the same production configurations which are needed to build accurate ML and DL prediction models.

	%In this study, we deployed IoT sensors in strawberry production polytunnels to monitor environmental conditions for strawberry yield forecasting. We also collected strawberry yield data before and after deploying sensors in polytunnel. To used strawberry yield data before sensors deployment, we propose a backcasting approach to generate synthetic sensor data for seasons with missing sensor data, using historical nearby Met Office weather station and polytunnel sensor data.
	
	The contributions of our work are as follows:  
	\begin{itemize}
		% \item We describe a real-world deployment of  IoT sensors in polytunnels for strawberry yield production to collect environmental parameters, including water usage, internal and external temperature, internal and external humidity, soil moisture and temperature, and photosynthetically active radiation and the resulting dataset and highlight the challenges associated with data collection in this setting. Furtehrmore, we clean, merge, and pre-process each sensor value for further prediction tasks.   
		\item We demonstrate a real-world implementation of IoT sensors in polytunnel strawberry production, collecting comprehensive environmental parameters including water usage, temperature (internal and external), soil conditions (moisture and temperature), and photosynthetically active radiation. We document the challenges encountered during deployment and data collection in real-world agricultural settings, providing insights for similar implementations in the future.
		%\item We integrate different IoT sensor data and preprocess it to utilize AI models.
		% \item The primary obstacle to implementing AI models in smart agriculture is the absence or inadequacy of historical data from the farming field. Therefore, we propose a backcasting methodology to generate synthetic sensor data for polytunnels by using  nearby historical weather stations and polytunnel sensors data. 
		\item We propose a solution to the limited historical sensor data by developing a backcasting methodology that generates synthetic polytunnel sensor data.
		
		%\item We also perform yield prediction of each polytunnel. 
		% \item In this study, we have collected strawberry yield data before and after sensor deployment. Therefore, we used our generated synthetic data for seasons with missing sensor data. Then, we perform strawberry yield forecasting using generated synthetic data and real sensor data.    
		\item We validate our approach by developing yield forecasting models using both the combined dataset and the real dataset alone. Our results show that models incorporating synthetic data strongly outperform those trained exclusively on the limited real dataset.
		
	\end{itemize}  
	
	To our knowledge, this is among the first studies to reconstruct missing growing-season IoT observations from external weather archives for downstream crop yield forecasting in operational polytunnel agriculture.
	
	\section{Related Work}
	To address the challenges of food shortages brought on by rapid population growth and the challenges of sustainable development, IoT is playing a crucial role in facilitating the transition from labor-intensive traditional agriculture to data-driven smart farming \cite{huo2024mapping}. Numerous studies have explored the integration of IoT in agriculture for real-time monitoring of environmental conditions \cite{priyadarshan2024digital, morchid2024intelligent, song2024design}. Benyezz et al. \cite{benyezza2023smart} presented an IoT-based platform for monitoring and controlling greenhouse climate and irrigation using sensors and a fusion system-based fuzzy logic strategy. 
	IoT sensors-based farm monitoring can help maintain agricultural production's sustainability with less effort and time \cite{rajak2023internet}. 
	
	Several researchers have studied forecasting strawberry yield by leveraging environmental sensors and external weather data using machine learning and deep learning models.  Lee et al. \cite{lee2020framework} design framework for strawberry fruit yield and ripening date prediction. Using a trigonometric model, they interpolated weather station temperature and humidity to get the polytunnel-specific temperature and humidity. Then, they performed yield prediction considering transformed temperature and humidity. Onoufriou et al. \cite{onoufriou2023premonition} proposed a deep learning model for strawberry yield forecasting that utilizes data collected from IoT sensors within polytunnels on a demonstrator farm rather than an actual farm, along with external weather data. However, our study focuses on a) improving strawberry yield predictions by simulating missing seasonal sensor observations in polytunnels using data from external weather stations and b) applying this approach to an operational, real-world farm. 
	
	Beddows et al. \cite{ beddows2024multi} proposed an expert-informed global-to-local model and an agent-based system \cite{beddows2026agent} for strawberry yield forecasting. They collected yield data from different polytunnel farms and used temperature readings from the ERA5 climate model. However, they did not use sensor data. Maskey et al. \cite{ maskey2019weather} presented a machine learning-based strawberry yield prediction using sensor data in a field, the nearest weather station, and agroclimatic indices. They collected one season of strawberry yield from February to June 2018. However, the farm site is not in the polytunnel.
	
	Despite the increasing maturity of IoT systems \cite{sinha2022recent}, agricultural data collection still requires improvements in both quantity and quality to develop AI-based applications that meet market demands. To design a data-driven decision system, AI models require substantial historical data for effective training. However, incomplete or short-term time series and tabular data in agricultural farming hinder the development of useful and generalizable AI models. Various synthetic data generation techniques have been proposed to increase the number of training samples for AI model training \cite{joshi2024synthetic, bandara2021improving}. Morales-García et al. \cite{ morales2023evaluation} generated a synthetic time series corpus of greenhouse temperature using a Generative Adversarial Network (GAN). Their synthetic data improves predictions of temperature in the greenhouse. Tai et al. \cite{tai2023using} generated synthesized multivariate agricultural time series data using GANs. Using both their generated and actual data, they trained deep learning models to predict future pest populations. They demonstrated that the synthetic data can effectively replicate real-world scenarios when insufficient real data is available. Several researchers have used the backcasting approach to retrieve missing data from previous years  \cite{twumasi2022machine,saghafian2018backcasting}. Saghafian et al.  \cite{saghafian2018backcasting} applied a backcasting approach to reconstruct past precipitation and enhance the length of the climate data in two stations of Iran's northwestern Urmia and Tabriz regions using statistical models. This study employed backcasting techniques to generate synthetic sensor data for seasons with missing observations. The generated sensor data was then incorporated into the yield forecasting model to improve predictions.

	While data augmentation techniques (such as noise injection or oversampling) improve model robustness, they do not reconstruct missing data points. Deep learning-based synthetic generation such as GANs generally perform best with larger training datasets, a condition rarely met in operational farms. In contrast, our backcasting approach bypasses these data volume constraints. Rather than generating data from latent noise (e.g., GAN), it leverages a continuous, highly reliable external data source (Met Office weather) to reconstruct the specific, localized micro-climate of the polytunnel, making it uniquely suited for environments with severe historical data scarcity.

	\section{ Materials and methods}
	\subsection{Data collection}
\begin{figure}[!t]
	\centering
	\includegraphics[width=\columnwidth]{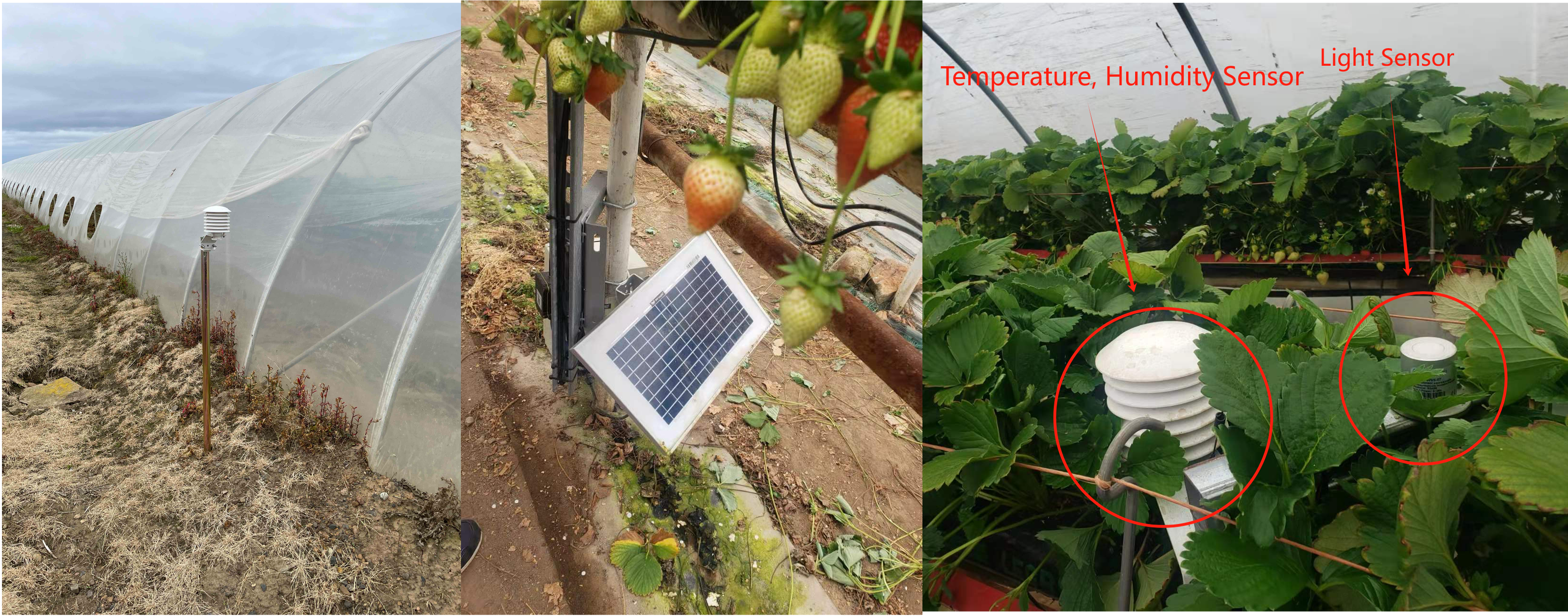}
	\caption{Example of our sensor deployment.}
	\label{fig:sensors}
\end{figure}

	This study was conducted in a polytunnel strawberry field located in Scotland, UK. IoT sensors were deployed to monitor environmental conditions inside and outside of Seaton and Multispan polytunnels over two growing seasons (an example can be seen in Figure \ref{fig:sensors}). Seaton is a heated polytunnel and contains four rows of tabletop strawberry plants. These strawberries were planted in January, and harvesting began in April and finished in mid-October.  Multispan is an unheated polytunnel that contains five rows of tabletop strawberry plants. Planting commenced in April, and harvesting started in May and ended in mid-October.
	
\begin{figure}[!t]
	\centering
	\begin{subfigure}[t]{\columnwidth}
		\centering
		\includegraphics[width=\columnwidth]{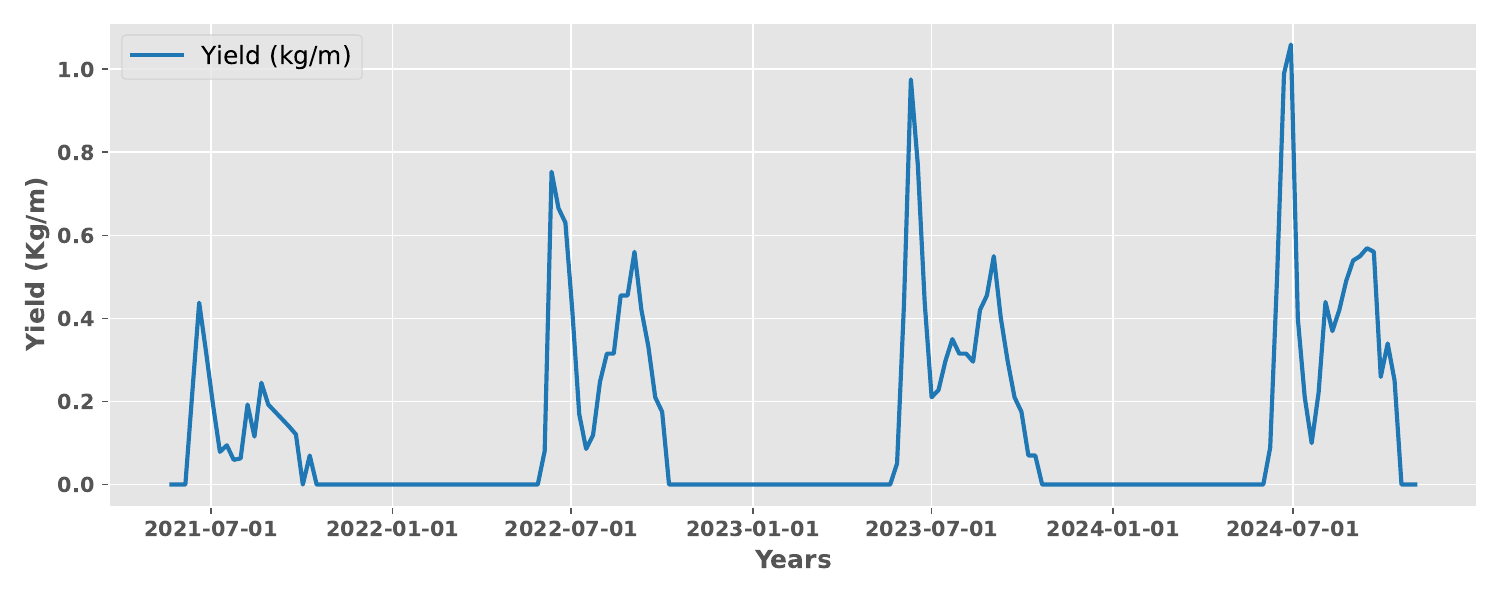}
		\caption{Four seasons strawberry yield for Multispan polytunnel.}
		\label{fig:yield_t9b}
	\end{subfigure}

	\vspace{0.5em}

	\begin{subfigure}[t]{\columnwidth}
		\centering
		\includegraphics[width=\columnwidth]{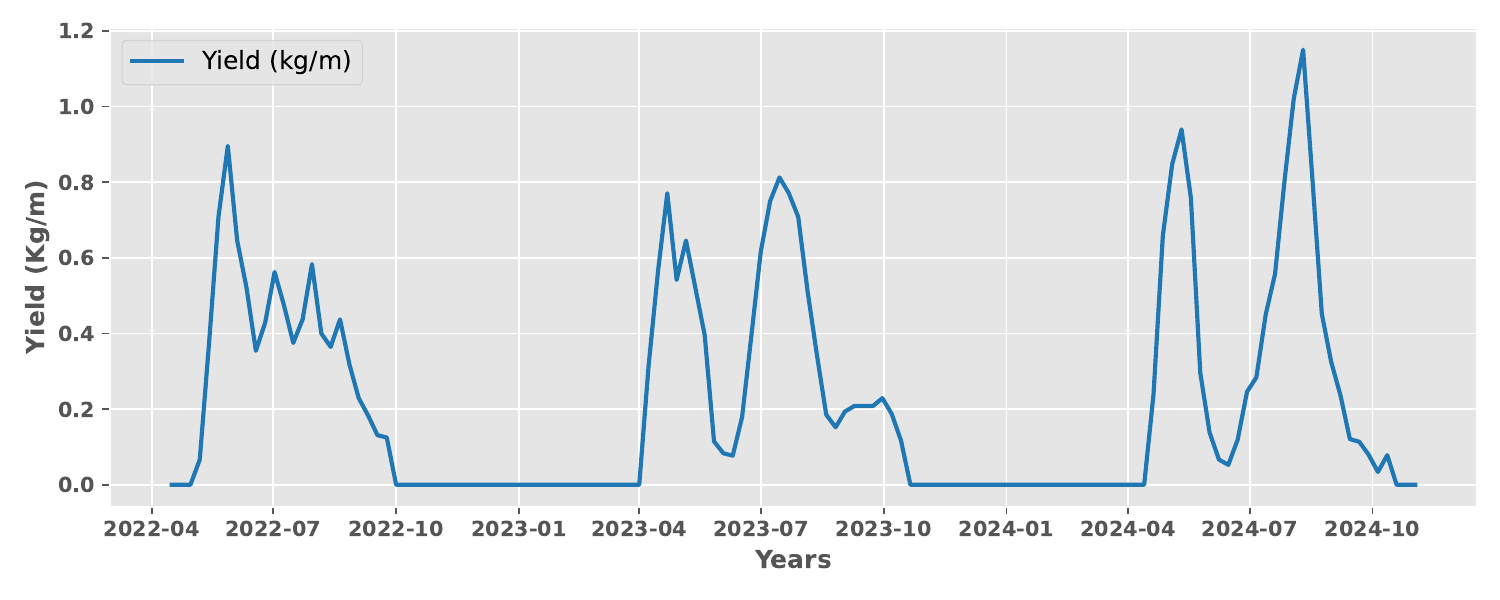}
		\caption{Three seasons strawberry yield for Seaton polytunnel.}
		\label{fig:yield_t45}
	\end{subfigure}

	\caption{Strawberry yields for Multispan and Seaton polytunnels.}
	\label{fig:yield_comparison}
\end{figure}

	Historical yield data were collected before and after sensor deployment in two polytunnels. Before deployment, two years of historical yield data were gathered from the Multispan polytunnel (2021–2022) and one year from the Seaton polytunnel (2022). After deploying sensors in both polytunnels, historical yield data and sensor data were collected from 2023 to 2024. Sensor data were recorded for two growing seasons: from May to mid-October 2023 and from April to mid-October 2024. In 2024, an additional polytunnel was added to each type. Figures \ref{fig:yield_t9b} and \ref{fig:yield_t45} depict the historical yields for the Multispan and Seaton polytunnels, respectively. Furthermore, hourly weather data from the UK Meteorological Office (Met Office), corresponding to the nearest station to the farm, were also incorporated. Table \ref{tab:featureSets} presents the polytunnel sensor and Met Office weather features along with their measurements.
	
	\begin{table*}[!ht]
		\caption{List of polytunnel sensors and Met Office weather features.}
		\resizebox{\textwidth}{!}{\begin{tabular}{|l|l|l|l|}
				\hline
				\textbf{Polytunnel Sensors} & \textbf{Unit} & \textbf{Met Office (MO) Weather} & \textbf{Unit} \\ \hline
				Water usage (WU) &      Litres      & Met Office external temperature (MET)& $^{\circ}C$    \\ \hline
				Internal temperature (IT) & $^{\circ}C$            & Visibility (Vis)& dm     \\ \hline
				Internal humidity (IH)&     \%        & Pressure (Pre)& hPa\\ \hline
				Polytunnel external temperature   (PET)&  $^{\circ}C$            & Met Office external humidity (MEH)& \%\\ \hline
				Polytunnel external humidity (PEH) &   \%            &  Radiation (Rad)  & KJ/m2\\ \hline
				Soil moisture (SM))&       \%      &  Wind speed (WS)& Kn  \\ \hline
				Soil temperature (ST) &    $^{\circ}C$         &Wind direction (WD) &  $^{\circ}$ \\ \hline
				Photosynthetically active radiation (PAR)&  $\mu$mol $m^{-2} s^{-1}$  & Wind gust (WG)& Kn \\ \hline
				Yield & Kg/m &  Rainfall (RFA)& mm \\ \hline
		\end{tabular}}
		\label{tab:featureSets}
	\end{table*}

	\subsection{Data preparation}
	\subsubsection{Data preprocessing}
	During deployment, the IoT sensors transmitted raw measurements at varying intervals ranging from 20 to 60 minutes. To establish a temporally consistent baseline across all environmental features, these readings were first synchronized and standardized into hourly averages.
	During data collection, we encountered various technical issues, such as power supply failures and environmental interference, which caused missing values. To address these gaps, we filled in missing PAR data using values from a nearby tunnel with similar sunlight exposure. For other sensors with intermittent missing data, we used the most recent available values.  In 2024, the WU sensor failed to send its recording value for a couple of months. Therefore, a forecasting model was developed to estimate the missing WU data using other sensors and MO data. %The results of this forecasting model are found in Figure \ref{fig:waterusage_multispan_seaton} and Table \ref{tab:waterusagepred}. 
	We deployed a single SM sensor in each tunnel in 2023; however, we observed spatial variability across locations within the polytunnel. Thus, we deployed three SM sensors in each tunnel in 2024 and took the average for a better estimate of the SM.
	After cleaning and preprocessing, we obtained 27,932 hourly data points from both tunnel types. Additionally, we applied One-Hot Encoding to differentiate the two tunnel types, Seaton and Multispan. 
	
	To increase the training data's volume, we used historical Met Office data to predict missing sensor values for earlier years. Specifically, while we had both yield and sensor data for 2023 and 2024, the volume of data was limited. However, we also had yield data for 2021 and 2022, although no sensor data was available since the sensors had not yet been deployed. Using a predictive model, we estimated sensor values for these years to obtain more training data. We split the dataset into 85\% for training and 15\% for testing, generating an additional 19,825 hourly data points. These synthetic data points were then combined with the real dataset to increase the training volume. Finally, since yield was reported weekly, we further downsampled the daily data to weekly intervals. \rev{This data augmentation should be interpreted as retrospective reconstruction of missing historical sensor observations rather than as a fully prospective deployment protocol.} 
	
	\subsubsection{Correlation analysis of features}
	A correlation analysis was conducted using Pearson Correlation ($p$) as expressed in Eqs. \ref{eg:pcc} to better understand the relationships between weather, polytunnel sensors, and yield features.
	\begin{equation}
		p = \frac{\sum_{i=1}^{N}(X_i-\bar{X})(Y_i-\bar{Y})}
		{\sqrt{\sum_{i=1}^{N}(X_i-\bar{X})^2}
			\sqrt{\sum_{i=1}^{N}(Y_i-\bar{Y})^2}},  
		\label{eg:pcc}
	\end{equation}
	where $N$ is the size of samples,  $X_i$ and $Y_i$ are the $i^{th}$, samples, $\bar X$ and $\bar Y$ are the mean values $X_i$ and $Y_i$, respectively. The value of  $p$ ranges from -1 to 1. The degree of correlation can be classified as \cite{schober2018correlation}, $0.00  < |p|\leq0.10$ is a negligible correlation, $0.10 < |p| \leq 0.39$ is a weak correlation, $0.39 < |p| \leq 0.69$ is a moderate correlation, $0.69 < |p| \leq0.89$ is a strong correlation and $0.89 < |p| \leq 1$ is a very strong correlation.    
	\subsubsection{Normalization}
	To reduce the influence of multidimensional features and speed up model convergence, the linear normalization method was used to keep data in the same dimension. We normalized all features to the same scale using min-max normalization as expressed in Eqs. \ref{eq:norm}. 
	\begin{equation}
		\label{eq:norm}
		x_{i} ^ {'} = (a_2 - a_1) \frac{x_{i} - min(x_i)}{max(x_i) - min(x_i)} + a_1 ,  
	\end{equation}
	where \rev{$a_1=-1$ and $a_2=1$ in this study; hence} $x_{i} ^ {'}$ is a normalized value \rev{with $x_i' \in [a_1,a_2]=[-1,1]$. Normalization was used during model fitting, while reported RMSE and MAE values are given in the original physical units of each variable after inverse transformation of predictions and observations.} 
	
	\subsection{Forecasting vs Backcasting approaches }
	Time series forecasting uses historical data to predict future trends or outcomes.  A given time series \{$\phi _t,.., \phi _{t-L}\}$, where $t$ is the value of each series at the time, and $L$ is the number of historical values to look back at. The forecasting model can be expressed as:
	\begin{equation}
		\hat{\phi}_{t+h} = f(\phi_{t},...,\phi_{t-L}),
	\end{equation}
	where $f$ is the forecasting function, $ \hat{\phi}_{t+h} $ is forecasting  at a time $t+h$, and $h$ is a forecasting horizon, which is $h\geq 1$.
	
	Time series backcasting is a complementary strategy to forecasting that uses observations available after a target time to reconstruct earlier, unobserved states. In this study, the retrospective backcasting task estimates a missing polytunnel sensor value from archived future Met Office covariates. For polytunnel type $c \in \mathcal{C}$ and target sensor $j \in \mathcal{J}$, the backcasting problem can be expressed as:
	\begin{equation}
		\label{eq:backcasting}
		\hat{s}_{c,t}^{(j)} =
		b_{c,j}(\mathbf{M}_{t+1}, \mathbf{M}_{t+2}, \ldots, \mathbf{M}_{t+k}),
	\end{equation}
	where $\hat{s}_{c,t}^{(j)}$ is the reconstructed sensor value and $b_{c,j}$ is the polytunnel- and sensor-specific backcasting model.
	\subsection{Proposed system}
\begin{figure}[!ht]
		\centering
		\includegraphics[scale=0.45]{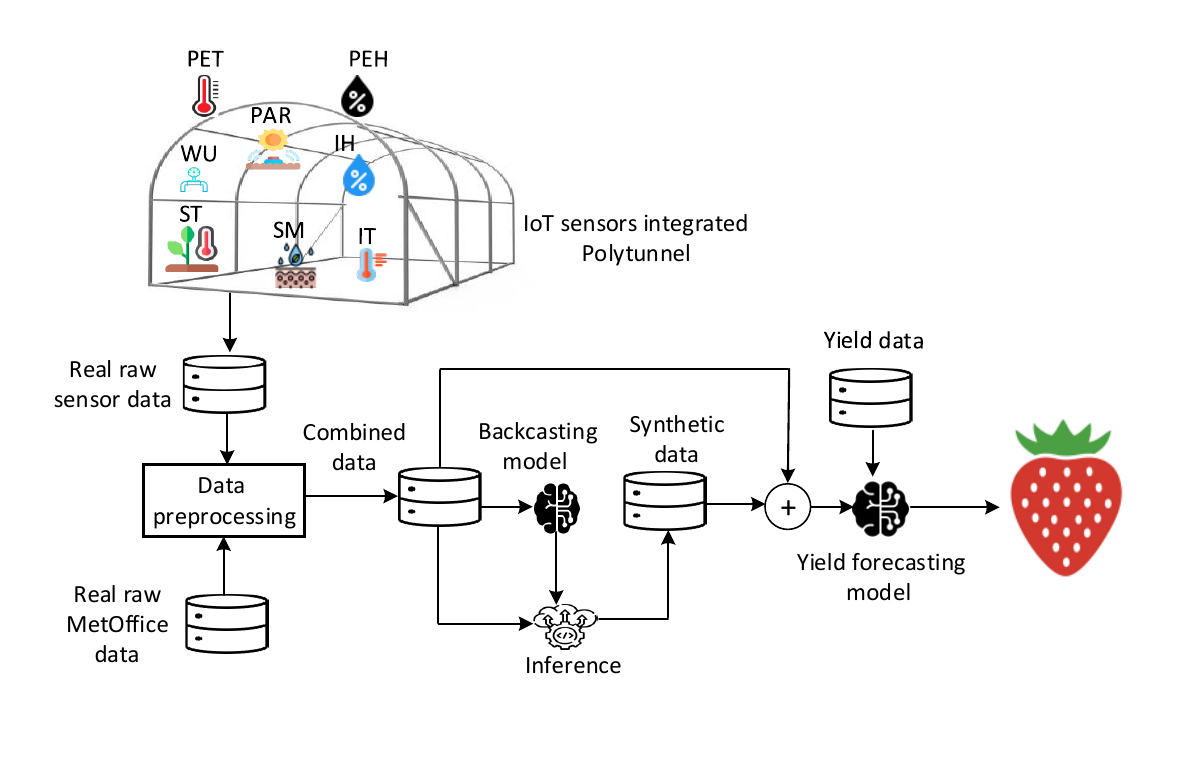}
		%\includegraphics[width=\textwidth]{Figures/Systemflow_recent_final.pdf}
		%\includepdf[pages=-]{Figures/Systemflow_recent_final.png}
		\caption{Schematic of our experimental setup and computational pipeline.}
		\label{fig:diagram}
	\end{figure}

	The proposed system comprises three key components: data collection, synthetic data generation, and yield forecasting. Figure \ref{fig:diagram} illustrates the pipeline of the proposed system. IoT sensors were integrated into strawberry production polytunnels for two growing seasons to collect environmental data, including WU, PET, IT, PEH, IH, SM, ST, and PAR. These IoT sensors and Met Office data were preprocessed before being applied to the downstream task.
	
	To address the gap of missing IoT observations for two additional seasons, a synthetic data generation method was proposed using a backcasting approach. The backcasting model was designed to replicate the environmental conditions within the polytunnel based on external meteorological observations.
	To train the models in the backcasting approach, real deployed-season Met Office and sensor data were aligned by timestamp. For each time index $t$, the observed sensor target was paired with a look-ahead window of archived Met Office observations at $t+1,\ldots,t+k$; throughout Algorithm \ref{alg1}, these indices denote later real-time observations relative to $t$. During inference, the best model was selected to generate synthetic data by inputting historical Met Office data from seasons without sensor deployment. Finally, a yield forecasting model was developed to combine real and synthetic IoT observations to evaluate the proposed approach.
	
	The proposed framework is intended for retrospective model training when historical yields are available but sensor deployments began later. Thus, future weather observations relative to the reconstructed season are already archived and legitimately available during reconstruction. \rev{Accordingly, the evaluation in this manuscript should be interpreted as a retrospective assessment of sensor-data augmentation under historical data scarcity, not as a fully prospective hold-out in which the backcasting model is also independent of the designated yield-evaluation year. The held-out 2023 yield labels are not used for yield-model training; however, the backcasting stage is calibrated using deployed-season sensor--weather relationships, including 2023 and 2024.}
	
	\subsection{Machine Learning Models}
	While deep learning and sequence-aware models (such as LSTMs \cite{hochreiter1997long} or Transformers \cite{vaswani2017attention, lim2021temporal, li2017novel}) are powerful for time-series data, they were inappropriate for our use case as they possess high parameter counts and require extensive datasets to train or fine-tune. For instance, Sun et al. trained a CNN-LSTM for soybean prediction using satellite images spanning across 5 years in 15 US states \cite{sun2019county}; Khaki et al. predicted soybean and corn yields using tens of thousands of tabular data points from 13 US States collected over 3 decades\cite{khaki2020cnn}. Huang et al. used transformers on 15 datasets, ranging from $\sim$1000 to $\sim$425,000 tabular rows, and found that using transformers with smaller datasets (e.g., under 10,000 rows) consistently gave worse performance than GBDT \cite{huang2020tabtransformer}. Tree-based methods have been shown to still outperform modern deep learning architectures and remain the most reliable machine learning technique to date for tabular data \cite{grinsztajn2022tree}. We face a very specific but pragmatic real-world problem, that individual farms simply do not have enough data for the most cutting-edge deep learning models. Because our target variable (yield) is aggregated weekly, our final dataset contains merely hundreds of temporal data points, which is far less than even a fraction of what is required to stably train a deep neural network.
	
	Under these data scarcity constraints, ensemble models \cite{yang2023survey} like Random Forest and XGBoost provide superior robustness, resist overfitting, and generalize better than deep learning architectures. Rather than relying on a single model, ensemble approaches leverage the strengths of multiple models to achieve better outcomes than any individual model. This technique is based on the notion that a group of weak learners can combine to generate a strong learner. Random forest (RF), Gradient Boosting Decision Tree (GBDT), and Extreme Gradient Boosting (XGBoost) are types of ensemble learning models that perform classification or regression tasks. RF uses a bagging method, which involves splitting the original dataset into several parts, training different models on each subset, and then combining their predictions. However, GBDT and XGBoost \cite{chen2016xgboost} use a boosting method for training data and a gradient descent optimizer to minimize the loss function. The boosting technique combines a weak learner sequentially to form a strong learner. Each following weak learner concentrates on the prior learner's mistakes, progressively reducing the prediction error. These models are popular machine learning techniques widely used for yield prediction \cite{beddows2024multi, shahhosseini2020forecasting}.
	For all ensemble models, hyperparameters were optimized using Grid Search with 3-fold cross-validation to ensure optimal configuration without overfitting on our limited sample size.
	
	\subsection{Algorithm}
\rev{Algorithm \ref{alg1} summarizes the sensor-wise direct backcasting procedure. For each polytunnel type $c$ and target sensor $j$, deployed-season sensor observations are paired with look-ahead windows of archived Met Office covariates. Each window $\mathbf{W}_t=[\mathbf{M}_{t+1},\ldots,\mathbf{M}_{t+k}]$ is converted to a tabular input vector $\mathbf{x}_t=\operatorname{vec}(\mathbf{W}_t)$ and used to learn a mapping from external meteorological conditions to the corresponding internal sensor value $s_{c,t}^{(j)}$. During inference, the selected model is applied to historical Met Office windows from seasons without sensor deployment. Sensor observations and yield labels are not used as inputs during this reconstruction.}

\begin{algorithm}[]
	{
		\caption{Sensor-wise Direct Backcasting from Met Office Data}
		\label{alg1}
		\KwIn{
			Deployed-season Met Office covariates $\mathbf{M}_t$ and sensor targets $s_{c,t}^{(j)}$\;
			Historical Met Office covariates $\mathbf{M}^{hist}_t$\;
			Candidate regression models $\mathcal{A}$ and window size $k$
		}
		\KwOut{
			Synthetic historical sensor series $\hat{S}_{c}^{(j)}$ and selected model $A_{c,j}^{*}$
		}
		\For{$c \in \mathcal{C}$}{
			\For{$j \in \mathcal{J}$}{
				Initialize $\mathcal{D}_{c,j} \leftarrow \emptyset$\;
				\For{each deployed-season $t$ with $t+k$ available}{
					$\mathbf{x}_t \leftarrow \operatorname{vec}([\mathbf{M}_{t+1},\ldots,\mathbf{M}_{t+k}])$\;
					Add $(\mathbf{x}_t,s_{c,t}^{(j)})$ to $\mathcal{D}_{c,j}$\;
				}
				Train and evaluate candidate models in $\mathcal{A}$ on $\mathcal{D}_{c,j}$\;
				Select $A_{c,j}^{*}$ with the lowest evaluation RMSE and set its fitted regressor as $f_{c,j}$\;
				Initialize $\hat{S}_{c}^{(j)} \leftarrow [\,]$\;
				\For{each historical $t$ with $t+k$ available}{
					$\mathbf{x}^{hist}_t \leftarrow \operatorname{vec}([\mathbf{M}^{hist}_{t+1},\ldots,\mathbf{M}^{hist}_{t+k}])$\;
					$\hat{s}_{c,t}^{(j)} \leftarrow f_{c,j}(\mathbf{x}^{hist}_t)$\;
					Append $(t,\hat{s}_{c,t}^{(j)})$ to $\hat{S}_{c}^{(j)}$\;
				}
			}
		}
		\Return{$\{\hat{S}_{c}^{(j)}, A_{c,j}^{*}: c \in \mathcal{C}, j \in \mathcal{J}\}$}\;
	}
\end{algorithm}
	
	\subsection{Evaluation metrics}
	The model accuracy is evaluated using Root Mean Squared Error (RMSE) and Mean Absolute Error (MAE). \rev{Although models were fitted using normalized variables, the reported error metrics are computed after inverse transformation to the original physical scale of each target variable. Thus, for example, PAR errors in Table \ref{Tab:backcasting} are reported in $\mu$mol $m^{-2} s^{-1}$ rather than on the normalized scale.} The errors can be calculated as follows:
	\begin{itemize}
		
		\item RMSE is the square root of the average of the squared difference between predicted and actual values. The RMSE is defined as: 
		\begin{equation}
			\label{Egs5}
			RMSE=\sqrt{{\frac{1}{N}\sum_{i=1}^{N} ({y_{i}^{'} - y_i})^2}}. 
		\end{equation}     
		
		\item MAE is the average of the absolute difference between predicted and actual values. The MAE is defined as: 
		\begin{equation}
			\label{Eqs6}
			MAE = {\frac{1}{N}\sum_{i=1}^{N} |{y_{i}^{'} - y_{i}}|},
		\end{equation}
		where $y_i'$ is the predicted, $y_i$ is the actual value, and $N$ is the total number of values in the test set. When evaluating multiple models, the model with the smallest RMSE and MAE performs better.
		
	\end{itemize}
	\section{Results and Discussion}
	\subsection{Correlation analysis of features}
\begin{figure*}[!ht]
		\centering
		\begin{subfigure}[b]{0.85\textwidth}
			\centering
			\includegraphics[width=\textwidth]{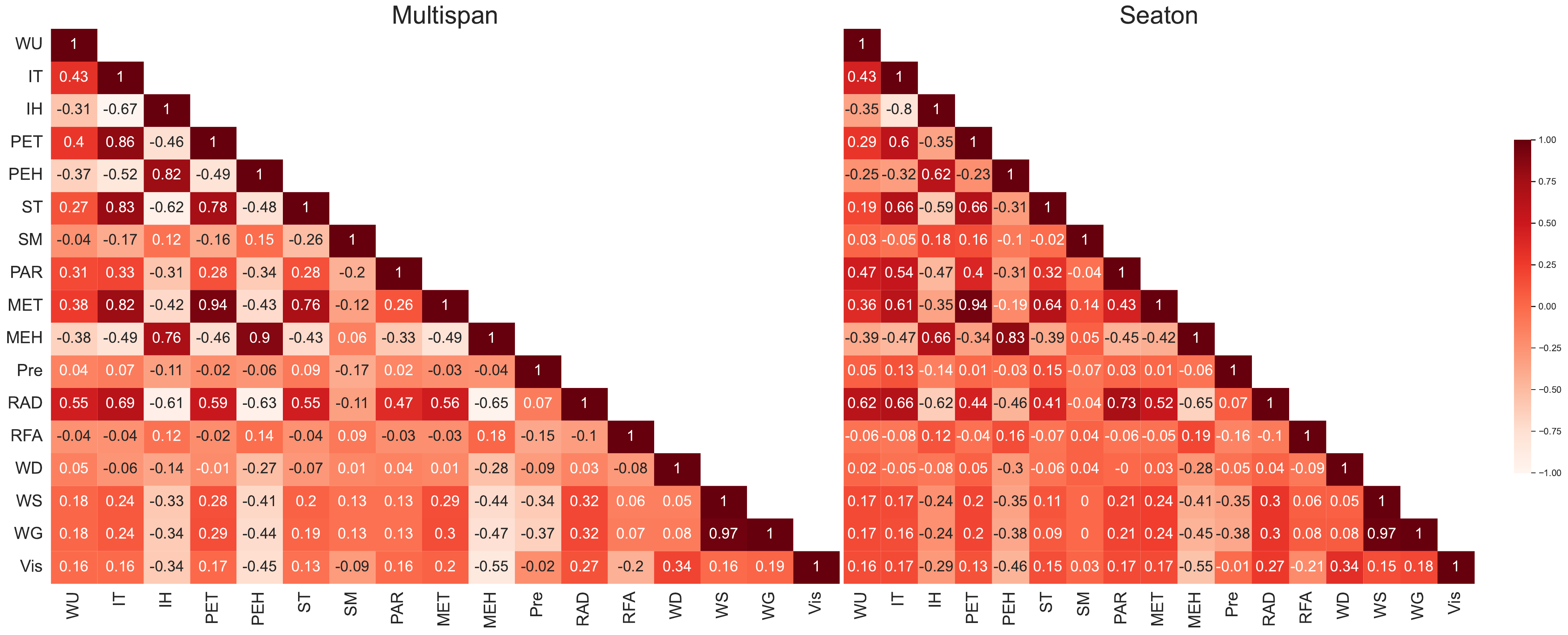}
			\caption{Hourly Pearson correlation of Multispan, and Seaton polytunnels sensor and Met Office weather data for the 2023 and 2024 seasons.        
				The left and right sides of the figure illustrate Multispan and Seaton polytunnels, respectively.}
			\label{fig:cor_multispan_seaton_weather}
		\end{subfigure} %
		\hfill % Add space between subfigures
		\begin{subfigure}[b]{0.85\textwidth}
			\centering
			\includegraphics[width=\textwidth]{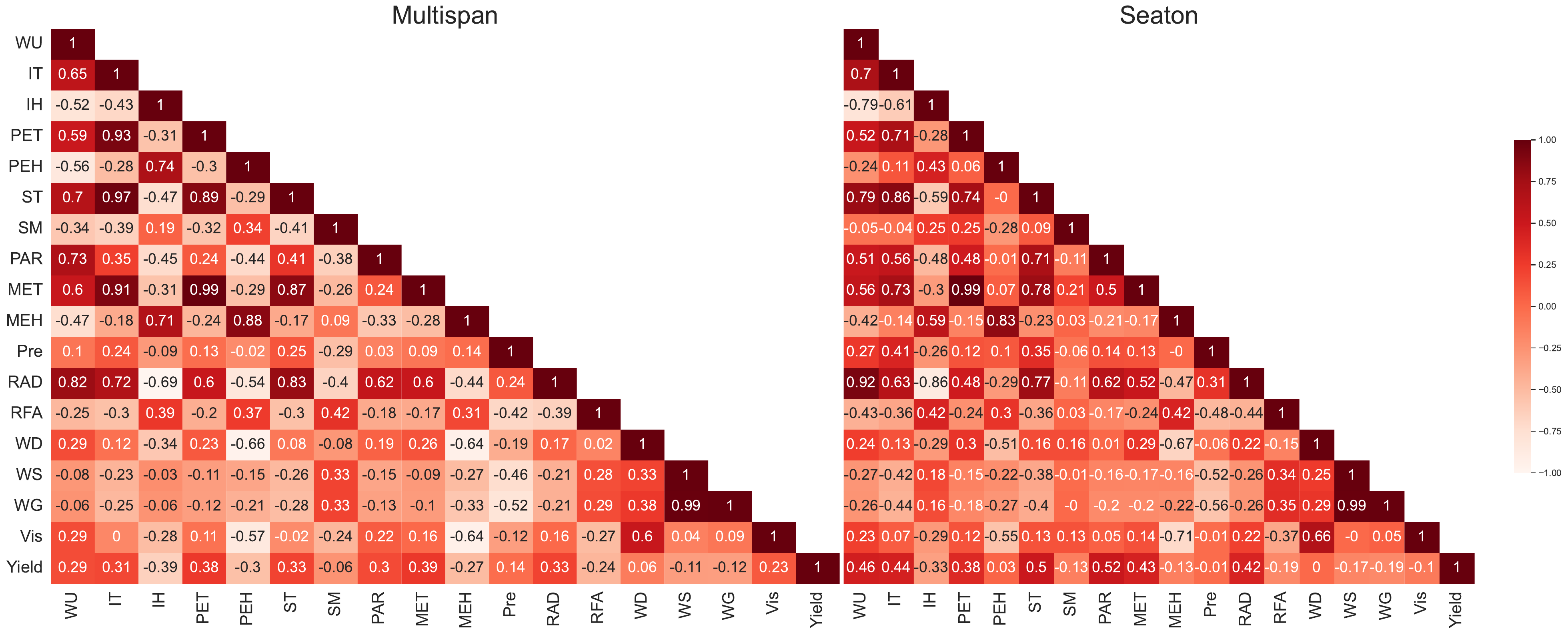}
			\caption{Weekly Pearson correlation of Multispan, and Seaton polytunnels sensor, Met Office weather, and Yield data for the 2023 and 2024 seasons. The left and right sides of the figure illustrate Multispan and Seaton polytunnels, respectively.}
			\label{fig:corr_multispan_seaton_yield}
		\end{subfigure}
		\caption{Pearson correlation analysis for Multispan and Seaton polytunnels.}
		\label{fig:pearson_cor}
	\end{figure*}

	We used Pearson Correlation to analyze the relationship between sensors, Met Office, and yield data. The correlation between hourly sensors and Met Office data was illustrated in Figure \ref{fig:cor_multispan_seaton_weather}. Readers can refer to Table \ref{tab:featureSets} for a detailed description of the features' acronyms. In both polytunnels, PEH and MEH exhibit a very strong correlation that indicates a close relationship. Strong correlations were observed between features such as IT and PET, IT and ST, PET and MET, and ST and MET in the Multispan polytunnel, and IT and ST, PET and MET, and PAR and Rad in the Seaton polytunnel, highlighting their interdependence. Moderate correlations include IT and MET, IT and Rad, and ST and Rad WU and IT, in both polytunnels, as well as PAR and MET in the Multispan polytunnel and WU and PAR in the Seaton polytunnel, suggesting moderate predictive relationships.
	Weak correlations are prominent for SM, Pre, WS, WG, and Vis across most features, indicating minimal influence on other variables. Based on the above correlation results,  we used Met Office data to generate polytunnel sensor data via the backcasting approach. The backcasting model used Met Office covariates only, since polytunnel sensor observations were unavailable during historical inference. For downstream yield forecasting, all features were used, with the external temperature pair (PET, MET) and humidity pair (PEH, MEH) replaced by their mean values because each pair was strongly correlated and had similar functionality.  
	
	The weekly correlation between yield and environmental parameters for the Multispan and Seaton polytunnels is illustrated in Figure \ref{fig:corr_multispan_seaton_yield}. The yield from the Multispan polytunnel exhibits a weak correlation with both sensor and Met Office features. Similarly, the yield from the Seaton polytunnel shows a moderate correlation with sensor features WU, IT, ST, PAR, and with Met Office features MET and Rad, and a weak correlation with other sensor and Met Office features.

	\subsection{Synthetic data generation}
	To generate synthetic data using our backcasting approach, we utilized a window size of six future values as input features for predicting past sensor readings. The input features were derived from Met Office observations (MET, MEH, Pre, Vis, Rad, WS, WD, WG,  and RFA). These features provide valuable information about the external environmental conditions surrounding the polytunnel. During model training, the objective was to predict past sensor values (WU, IT, IH, SM, ST, and PAR). Since sensor data is unavailable during inference (i.e., when predicting unseen historical values), sensor observations from future time steps were intentionally excluded from the model's input features. Instead, the model relied solely on the Met Office to infer the internal sensor conditions. By learning these patterns during training, the model can reconstruct past sensor values without direct access to future sensor readings.
	
	\rev{The backcasting stage therefore reconstructs sensor variables from external meteorological covariates, and it does not use yield labels. However, because the backcasting calibration uses sensor--weather relationships from the deployed sensor seasons, its outputs should be interpreted as retrospective reconstructions rather than fully prospective predictions.}
	
	We utilized actual data from 2023 and 2024 for both the training and testing phases of the models. Subsequently, we generated two years of synthetic sensor data by inputting MO data from 2021 and 2022 into the models. Table \ref{Tab:backcasting}  presents the RMSE and MAE of the RF, GBDT, and XGBoost models for backcasting  Multispan and Seaton polytunnel features. The model with the lowest RMSE of each feature was selected to generate synthetic data. For the Multispan polytunnel, the RF model was selected to generate synthetic data for WU, IT, IH, and PAR, while the GBDT model was chosen to generate ST and SM. For the Seaton polytunnel, the RF model was used to generate WU, the XGBoost model was selected to generate SM, ST, and PAR, and the GBDT model was used to generate IT and IH. 
	%The actual and backcasted values of each polytunnel feature are illustrated in Figures \ref{fig:backWU} to \ref{fig:backPAR} in the appendix. 

	\begin{table*}
		\centering
		\caption{\rev{Polytunnel sensor backcasting model performance. RMSE and MAE are reported after inverse transformation in the original physical units of each sensor. Bold RMSE values denote the lowest RMSE for each feature and polytunnel.}}
		\begin{tabular}{|l|l|l|llllll|}
			\hline
			\multirow{2}{*}{Models} & \multirow{2}{*}{Polytunnels} & \multirow{2}{*}{Metrics} & \multicolumn{6}{l|}{Features}                                                                                                                              \\ \cline{4-9} 
			&                              &                           & \multicolumn{1}{l|}{WU}     & \multicolumn{1}{l|}{IT}   & \multicolumn{1}{l|}{IH}    & \multicolumn{1}{l|}{ST}    & \multicolumn{1}{l|}{SM}    & PAR     \\ \hline
			\multirow{4}{*}{RF}     & \multirow{2}{*}{Multispan}   & RMSE                      & \multicolumn{1}{l|}{\textbf{11.118}} & \multicolumn{1}{l|}{\textbf{3.111}} & \multicolumn{1}{l|}{\textbf{7.839}} & \multicolumn{1}{l|}{3.1} & \multicolumn{1}{l|}{4.11} & \textbf{308.847} \\ \cline{3-9} 
			&                              & MAE                       & \multicolumn{1}{l|}{8.074}  & \multicolumn{1}{l|}{2.253} & \multicolumn{1}{l|}{5.656}  & \multicolumn{1}{l|}{2.239} & \multicolumn{1}{l|}{3.825} & 203.929 \\ \cline{2-9} 
			& \multirow{2}{*}{Seaton}      & RMSE                      & \multicolumn{1}{l|}{\textbf{11.764}}       & \multicolumn{1}{l|}{3.984}      & \multicolumn{1}{l|}{10.097}       & \multicolumn{1}{l|}{2.933}      & \multicolumn{1}{l|}{ 5.593}      & 304.642        \\ \cline{3-9} 
			&                              & MAE                       & \multicolumn{1}{l|}{8.717}       & \multicolumn{1}{l|}{2.895}      & \multicolumn{1}{l|}{7.719}       & \multicolumn{1}{l|}{2.221}      & \multicolumn{1}{l|}{4.339}      & 166.146        \\ \hline
			
			\multirow{4}{*}{GBDT}     & \multirow{2}{*}{Multispan}   & RMSE                      & \multicolumn{1}{l|}{11.214} & \multicolumn{1}{l|}{3.145} & \multicolumn{1}{l|}{7.877} & \multicolumn{1}{l|}{\textbf{2.866}} & \multicolumn{1}{l|}{\textbf{4.062}} &315.633  \\ \cline{3-9} 
			&                              & MAE                       & \multicolumn{1}{l|}{7.869}  & \multicolumn{1}{l|}{2.223} & \multicolumn{1}{l|}{5.514}  & \multicolumn{1}{l|}{2.01} & \multicolumn{1}{l|}{3.702} & 208.180 \\ \cline{2-9} 
			& \multirow{2}{*}{Seaton}      & RMSE                      & \multicolumn{1}{l|}{11.866}       & \multicolumn{1}{l|}{\textbf{3.807}}      & \multicolumn{1}{l|}{\textbf{8.941}}       & \multicolumn{1}{l|}{2.459}      & \multicolumn{1}{l|}{5.437}      & 308.644        \\ \cline{3-9} 
			&                              & MAE                       & \multicolumn{1}{l|}{8.651}       & \multicolumn{1}{l|}{2.782}      & \multicolumn{1}{l|}{6.573}       & \multicolumn{1}{l|}{1.852}      & \multicolumn{1}{l|}{4.185}      &171.204        \\ \hline
			
			\multirow{4}{*}{XGBoost}     & \multirow{2}{*}{Multispan}   & RMSE                       & \multicolumn{1}{l|}{11.256}       & \multicolumn{1}{l|}{3.184}      & \multicolumn{1}{l|}{8.028}       & \multicolumn{1}{l|}{2.885}      & \multicolumn{1}{l|}{4.112}      &323.877        \\ \cline{3-9}  
			&                              & MAE                       & \multicolumn{1}{l|}{7.906}  & \multicolumn{1}{l|}{2.257} & \multicolumn{1}{l|}{5.649}  & \multicolumn{1}{l|}{2.031} & \multicolumn{1}{l|}{3.722} &  212.593\\ \cline{2-9} 
			& \multirow{2}{*}{Seaton}      & RMSE                      & \multicolumn{1}{l|}{11.842}       & \multicolumn{1}{l|}{3.832}      & \multicolumn{1}{l|}{9.073}       & \multicolumn{1}{l|}{\textbf{2.447}}      & \multicolumn{1}{l|}{\textbf{5.354}}      & \textbf{301.045}        \\ \cline{3-9} 
			&                              & MAE                       & \multicolumn{1}{l|}{8.632}       & \multicolumn{1}{l|}{ 2.770}      & \multicolumn{1}{l|}{6.641}       & \multicolumn{1}{l|}{1.862}      & \multicolumn{1}{l|}{4.101}      & 168.774         \\ \hline
			
		\end{tabular}
		\label{Tab:backcasting}
	\end{table*}
	
	\subsection{Yield forecasting}
	\subsubsection{Experiment setup}
	We designed our experiments to evaluate two key aspects: 1) the effectiveness of artificially enhancing the data volume with synthetic data, and 2) the impact of incorporating environmental features alongside the historical yield data for yield prediction. We processed the sensor data and MO data into weekly averages while aggregating yields into weekly sums. Given that yield forecasting is dependent on historical data from previous weeks, we processed data into sliding windows containing data for three consecutive weeks. Specifically, it uses environmental features from weeks 1-3, and yield from weeks 1 and 2 to predict the yield of the third week. After data cleaning and processing, we have 167 weekly samples of real data, and 81 weekly samples of synthetic data (note only sensor data are synthetic; yield and Met Office are real historical values).
	\rev{To approximate a practical retrospective deployment scenario,} we trained on data from all tunnels except those being tested (Seaton and Multispan from 2023). \rev{This split isolates the 2023 yield records for yield-model testing, but it should not be interpreted as making the backcasting calibration fully independent of 2023 environmental sensor data.}
	
	As with synthetic data generation, we evaluated RF, GBDT, and XGBoost. For each model, we tested four feature combinations - yield only, where only historical yield from weeks is used; yield + Met Office, where historical yield and Met Office data are used; Yield + Sensor, where historical yield plus sensor data in polytunnels is used; and yield + sensor + Met Office, where all features are used.

	\begin{table*}
		\centering
		\caption{Comparison of model performance using real data and data enhanced with synthetic data, validated on 2023 data. Values in bold represent the best (lowest) metrics for each model, with percentages showing performance improvement over the real and yield-only baseline.}
		\label{tab:yield-prediction-comparison}
		\resizebox{\textwidth}{!}{%
			\begin{tabular}{|c|c|cc|cc|cc|cc|}
				\hline
				\multirow{3}{*}{\textbf{Model}} & \multirow{3}{*}{\textbf{Data}} & \multicolumn{8}{c|}{\textbf{Feature Sets}} \\
				\cline{3-10}
				& & \multicolumn{2}{c|}{\textbf{Yield Only}} & \multicolumn{2}{c|}{\textbf{Yield + MO}} & \multicolumn{2}{c|}{\textbf{Yield + Sensor}} & \multicolumn{2}{c|}{\textbf{Yield + Sensor + MO}} \\
				\cline{3-10}
				& & \textbf{RMSE} & \textbf{MAE} & \textbf{RMSE} & \textbf{MAE} & \textbf{RMSE} & \textbf{MAE} & \textbf{RMSE} & \textbf{MAE} \\
				\hline
				\multirow{2}{*}{RF} & Real only & 0.1464 & \textbf{0.0940} & 0.1675 & 0.1140 & 0.1668 & 0.1137 & 0.1658 & 0.1147 \\
				& Syn+real & 0.1321 & 0.0957 & 0.1404 & 0.1088 & {\textbf{0.1260}$^{\color{blue}\uparrow 13.93\%}$} & 0.0961 & 0.1363 & 0.1049 \\
				\hline
				\multirow{2}{*}{GBDT} & Real only & 0.1514 & 0.1032 & 0.1801 & 0.1321 & 0.2098 & 0.1446 & 0.1692 & 0.1259 \\
				& Syn+real & 0.1418 & 0.1105 & 0.1435 & 0.1097 & {\textbf{0.1218}$^{\color{blue}\uparrow 19.55\%}$} & {\textbf{0.0952}$^{\color{blue}\uparrow 7.75\%}$} & 0.1352 & 0.1064 \\
				\hline
				\multirow{2}{*}{XGBoost} & Real only & 0.1571 & 0.1137 & 0.1869 & 0.1360 & 0.1934 & 0.1397 & 0.1691 & 0.1190 \\
				& Syn+real & 0.1314 & 0.1016 & 0.1468 & 0.1141 & {\textbf{0.1288}$^{\color{blue}\uparrow 18.01\%}$ } & {\textbf{0.0996}$^{\color{blue}\uparrow 12.40\%}$}& 0.1340 & 0.1017 \\
				\hline
			\end{tabular}
		}
	\end{table*}
	% \usepackage{multirow}
	% \usepackage{multirow}
	
	\subsubsection{Effect of data volume and environmental features}
	We established two key baselines for comparison: yield-only models to assess the impact of adding environmental features, and real-only data models to assess the value of synthetic sensor data. \rev{The yield-only setting is an autoregressive machine-learning baseline because it uses previous weekly yield values to predict the target week without environmental covariates.} The results are displayed in Table \ref{tab:yield-prediction-comparison}.
	
	% \paragraph{Impact of environmental features (yield-only baseline)}
	\noindent\textbf{Impact of environmental features (yield-only baseline):} Using the yield data alone provided reasonable prediction performance, with MAE of around 0.09 to 0.11, and RMSE of around 0.14 to 0.16. Considering that strawberry yields are typically between 0 to 1.2 kg/m, our results suggest that on average, predictions deviate from actual values by about 10\%. RMSE, which penalizes larger errors more than MAE due to squaring before averaging, can be more informative for evaluating peaks of the growing season, where yield variations are significant. 
	
	Interestingly, the models appear to be highly sensitive to the yield from previous weeks and less sensitive to the environmental features. When using only the real data, adding environmental features degraded performance across all models. This suggests insufficient data to learn complex environmental relationships. This finding aligns with established machine learning principles, which suggest that complex feature interactions require substantial training volume, a factor that appears to be the bottleneck in our case. \rev{Although the yield-only baseline captures autoregressive information, we did not include the simpler naive persistence rule $\hat{y}_t=y_{t-1}$. Therefore, we avoid claiming that the models outperform the simplest possible carry-forward predictor.}
	
	% \paragraph{Effect of boosting data volume with synthetic data (real-only baseline)}
	\noindent\textbf{Effect of boosting data volume with synthetic data (real-only baseline):} By augmenting the real dataset with synthetic samples, we observed a significant improvement across all models. As the models struggle to utilise the complexity of features due to the small training volume, we reduced feature complexity by selecting only the most yield-correlated sensor features (IT, IH, SM, PAR) and removing the less impactful ones (WU, PET, PEH, ST). We observed an improvement of 24.5\%, 41.9\%, and 33.4\% for RF, GBDT, and XGBoost respectively. Even for combinations not using our synthetic sensor data (yield-only and yield+MO), the additional historical yield and MO data from 2021-2022 substantially improved performance. This further confirms the importance of data volume. While yield+sensor+MO combinations also showed significant improvement over their real-only baselines, their overall performance remained slightly worse than yield+sensor combinations. This could be due to two reasons - first, there exists an optimal balance between feature richness and data volume that our current dataset approaches but does not fully achieve; and second, polytunnel sensor data likely provides more accurate measurements of the microenvironment inside the tunnels compared to regional meteorological data, which may not capture the controlled conditions within the polytunnels as effectively. \rev{We therefore interpret the improved Yield+Sensor results as evidence of practical predictive utility under data augmentation, rather than as proof that the models have learned causal agronomic mechanisms. Feature-importance or SHAP-style analyses would be needed to quantify the contribution of individual environmental covariates.}

	\begin{table}[]
		
		\centering
		\renewcommand{\arraystretch}{1.4} % increases row height (vertical spacing)
		\setlength{\tabcolsep}{10pt}      % increases column spacing (horizontal padding)
		\caption{Comparison of model yield prediction performance across different window sizes used for synthetic data generation. The data is syn+real to show the effect on yield prediction.}
		\begin{tabular}{|l|l|ll|}
			\hline
			\multirow{3}{*}{\textbf{Models}} & \multirow{3}{*}{\textbf{Window Sizes}} & \multicolumn{2}{l|}{\textbf{Feature set}} \\ \cline{3-4} 
			&                                        & \multicolumn{2}{l|}{\textbf{Yield+Sensor}}        \\ \cline{3-4} 
			&                                        & \multicolumn{1}{l|}{\textbf{RMSE}}      & \textbf{MAE}      \\ \hline
			\multirow{4}{*}{RF}              & 2                                      & \multicolumn{1}{l|}{0.1285}    & 0.0994   \\ \cline{2-4} 
			& 4                                      & \multicolumn{1}{l|}{0.1283}    & 0.0982   \\ \cline{2-4} 
			& 6                                      & \multicolumn{1}{l|}{\textbf{0.1260}}    & 0.0961   \\ \cline{2-4} 
			& 8                                      & \multicolumn{1}{l|}{0.1270}    & \textbf{0.0960}   \\ \hline
			\multirow{4}{*}{GBDT}            & 2                                      & \multicolumn{1}{l|}{0.1267}    & 0.0985   \\ \cline{2-4} 
			& 4                                      & \multicolumn{1}{l|}{0.1221}    & 0.0979   \\ \cline{2-4} 
			& 6                                      & \multicolumn{1}{l|}{\textbf{0.1218}}    & \textbf{0.0952}   \\ \cline{2-4} 
			& 8                                      & \multicolumn{1}{l|}{0.1220}    & 0.0995   \\ \hline
			\multirow{4}{*}{XGBoost}         & 2                                      & \multicolumn{1}{l|}{0.1431}    & 0.1134   \\ \cline{2-4} 
			& 4                                      & \multicolumn{1}{l|}{0.1375}    & 0.1067   \\ \cline{2-4} 
			& 6                                      & \multicolumn{1}{l|}{\textbf{0.1288}}    & \textbf{0.0996}   \\ \cline{2-4} 
			& 8                                      & \multicolumn{1}{l|}{0.1354}    & 0.1071   \\ \hline
		\end{tabular}
		\label{tab:windowsize}
	\end{table}
	\subsubsection{Effect of window size in synthetic data generation }
	
	The results presented in Table \ref{tab:windowsize} provide a comparative analysis of model performance across different window sizes used for synthetic data generation. A consistent pattern is observed across all algorithms, where a window size of 6 yields the lowest RMSE and near-lowest MAE values, indicating superior overall predictive accuracy. Specifically, RF, GBDT, and XGBoost achieve their best RMSE performance at this configuration, suggesting that this window size effectively captures the underlying temporal dependencies in the data.
	
	In contrast, smaller window sizes (2 and 4) result in relatively higher error values, which can be attributed to insufficient temporal context, limiting the models’ ability to learn meaningful sequential patterns. On the other hand, increasing the window size to 8 does not lead to further improvement; instead, a slight degradation in performance is observed. This may be due to the introduction of redundant or less relevant historical information, which can increase noise and reduce model generalization.
	
	Among the evaluated models, GBDT demonstrates the best overall performance, achieving the lowest RMSE (0.1218) and MAE (0.0952), followed closely by RF, while XGBoost exhibits comparatively higher error values across all window sizes. Overall, the findings highlight the critical role of window size selection in synthetic data-driven modeling and emphasize that an intermediate window size provides the best trade-off between information richness and noise reduction.
	
	\subsection{Limitations and Future Work}
	Despite significant predictive improvements, we acknowledge several limitations. First, training downstream models on partially synthetic data introduces the risk of error propagation; inaccuracies generated during backcasting may amplify errors in the final yield forecast. Second, evaluating on a single Scottish farm limits generalizability, and future work must validate transferability across diverse climates and polytunnel designs. Third, real-world IoT deployments are inherently vulnerable to hardware failures, power interruptions, and sensor dropouts. While our backcasting methodology successfully mitigates data issues caused by these hardware limitations, deploying this system in a live, operational setting will require real-time adaptability to handle unexpected sensor failures during active forecasting. \rev{Fourth, the present evaluation is retrospective: although the 2023 yield labels are held out from yield-model training, the backcasting model is calibrated using deployed-season sensor--weather relationships that include 2023. A stricter prospective evaluation would recalibrate the backcasting model without any sensor data from the designated yield-evaluation year. Fifth, we include an autoregressive machine-learning yield-only baseline, but we do not include a naive persistence baseline.} Finally, future research should explore advanced synthetic generation techniques capable of capturing extreme weather anomalies and their impacts on yield.

	\section{Conclusion}
	In this study, we deployed IoT sensors in polytunnel-based strawberry production to monitor environmental parameters and improve yield forecasting. To address the challenge of limited historical sensor data, we proposed a backcasting approach that generated synthetic sensor data using historical weather station and polytunnel data. Our results showed \rev{in a retrospective evaluation} that incorporating synthetic data improved yield forecasting accuracy, with models trained on the combined data outperforming those trained on real sensor data. We observed that incorporating additional environmental features did not improve accuracy when using only the limited real dataset. However, when combined with the synthetic data, the value of these environmental features became more pronounced, with the yield+sensor combination achieving the lowest RMSE across the three model families. \rev{These findings should be interpreted as evidence for retrospective data augmentation under sensor-data scarcity, while fully prospective validation and comparison against a naive persistence baseline remain important future work.}
	
	Our novel work addresses a practical challenge faced by individual farms: limited historical sensor data for predictive modeling. By combining IoT sensor deployment with synthetic data generation, we provide an approach that can be implemented in similar small-scale agricultural settings where data scarcity is common. 
	% \textcolor{blue}{While we achieved significant improvements, we acknowledge the limitations of using data from a single farm. Future work should explore transferability across multiple growing locations and other crop varieties. Further exploration should also be made into advanced synthetic data generation techniques that can capture extreme weather conditions and their effects on yield.}\\
	
	\noindent\textbf{Acknowledgments:} This work was supported by the UK Engineering and Physical Sciences Research Council, grant number EP/V042270/1.

		\bibliographystyle{IEEEtran} 
		\bibliography{biblio}
		
	\end{document}